%% file: main.tex
\documentclass[journal,twoside,web]{ieeecolor}
\usepackage{lcsys}
\usepackage{cite}
\usepackage{amsmath,amssymb,amsfonts}
\usepackage{algorithmic}
\usepackage{graphicx}
\usepackage{textcomp}
\usepackage{xcolor}
\usepackage[hidelinks]{hyperref}

\input{macros}
\def\BibTeX{{\rm B\kern-.05em{\sc i\kern-.025em b}\kern-.08em
    T\kern-.1667em\lower.7ex\hbox{E}\kern-.125emX}}
\begin{document}
\title{Providing Safety Assurances for Systems with Unknown Dynamics}
\author{Hao Wang, Javier Borquez, \IEEEmembership{Student Member, IEEE}, and Somil Bansal, \IEEEmembership{Member, IEEE}
\thanks{Manuscript received March 8, 2024; revised May 5, 2024; accepted May 21 2024. This work is supported in part by the NSF CAREER Program under award 2240163, the DARPA ANSR program, a Ming Hsieh Department of Electrical and Computer Engineering Fellowship. (Corresponding author: Hao Wang)}
\thanks{The authors are associated with the Ming Hsieh Department of Electrical and Computer Engineering, University of Southern California, CA 90089, USA. 
(emails: {\tt\small \{haowwang, javierbo, somilban\}@usc.edu})}
}
\pagestyle{empty}

\maketitle
\thispagestyle{empty}

\begin{abstract}
As autonomous systems become more complex and integral in our society, the need to accurately model and safely control these systems has increased significantly. In the past decade, there has been tremendous success in using deep learning techniques to model and control systems that are difficult to model using first principles. However, providing safety assurances for such systems remains difficult, partially due to the uncertainty in the learned model. In this work, we aim to provide safety assurances for systems whose dynamics are not readily derived from first principles and, hence, are more advantageous to be learned using deep learning techniques. Given the system of interest and safety constraints, we learn an ensemble model of the system dynamics from data. Leveraging ensemble uncertainty as a measure of uncertainty in the learned dynamics model, we compute a maximal robust control invariant set, starting from which the system is guaranteed to satisfy the safety constraints under the condition that realized model uncertainties are contained in the predefined set of admissible model uncertainty. We demonstrate the effectiveness of our method using a simulated case study with an inverted pendulum and a hardware experiment with a TurtleBot. The experiments show that our method robustifies the control actions of the system against model uncertainty and generates safe behaviors without being overly restrictive. The codes and accompanying videos can be found on the project website \footnote{\url{https://github.com/haowwang/safety_assurances_for_unknown_dynamics}}.
\end{abstract}


\section{Introduction}


\IEEEPARstart{A}UTONOMOUS systems are playing increasingly important roles in the functioning of modern society. However, traditional modeling techniques, such as using first principles, struggle to model these systems, given their increasing complexities. Recent advances in deep learning have enabled the modeling and control of systems with highly complex dynamics. While the methods have demonstrated strong control performance for a number of autonomous systems, they can lead to unsafe behaviors or even catastrophic failures due to the predictive uncertainty in the neural network models. 

In this work, we are interested in providing safety assurances for systems whose dynamics are unknown and difficult to model using first principles. Despite success in safety analysis for models developed from first principles, it remains difficult to provide safety assurances for systems with unknown or uncertain dynamics. Many works have utilized safety analysis frameworks, such as Control Barrier Function (CBF) \cite{ames17cbfqp} and Hamilton-Jacobi (HJ) reachability analysis \cite{bansal17hjsurvey}, to provide safety assurances to systems with unknown or uncertain dynamics. One popular line of works seeks to reduce model uncertainties to uncertainties in the CBF constraints, which are used to synthesize safety-critical controls \cite{Castaneda21,taylor20,niu2021cbf_safe_ctrl}. However, these methods rely on prior knowledge of the system to construct the CBFs, and more critically assume the CBFs constructed for the nominal model is valid for the actual system. This assumption can be easily violated when the model uncertainty is severe, and it is difficult to determine when the assumption is no longer valid. Another family of approaches is to generate safety guarantees for systems under the \emph{worst-case} model uncertainty using HJ reachability analysis. This line of work is rooted in differential game theory and its connection with the Hamilton-Jacobi-Isaacs partial differential equation (HJI PDE) \cite{barron1984viscosity, evans1984differential}. A time-dependent HJI PDE \cite{mitchell05} is formulated to study the pursuer-evader game, a type of two-person zero-sum differential game integral to HJ reachability analysis, and this formulation has been applied to generate safety assurances for system under dynamics uncertainties. More specifically, the authors in \cite{fisac19, herbert2021scalable, akametalu2014reach_gp_safe_learning} represent the model uncertainty as a Gaussian Process (GP) and utilize the predictive uncertainty of the GP to generate robust safety assurances online. However, the model uncertainties considered in the works depend only on the state and not the control input. Moreover, it is not immediately clear how similar techniques can be used for neural network-based dynamics models that does not predict uncertainties.

In its core, our method attempts to provide robust safety assurances for systems with unknown dynamics against model uncertainties. We first learn a nominal dynamics model of the system along with a measure of model uncertainty. Then, we compute robust safety assurances utilizing results from HJ reachability analysis. More specifically, given a set of undesirable states, our method computes the maximal robust control invariant set and a controller that renders the set control invariant for the model under bounded model uncertainty. A critical aspect of our work is that model uncertainty is both \emph{state and control-dependent}. This control dependence of model uncertainty is often ignored in prior works for simplicity. Specifically, when the model uncertainty is control-independent, the players in the two-person game do not interact directly and renders the game much easier to solve. Instead, we take this interplay between the control and model uncertainty into account and systematically handle that under the HJ reachability framework. This leads to a significantly less conservative estimate of the safe set of the system. Furthermore, we pay special attention to learned neural network nominal models. Using an ensemble of neural networks with control-affine architectures as the nominal model, we harness the modeling power of neural networks and obtain a heuristic measure of model uncertainty from the ensemble, with which we provide robust safety assurances for the system.

Our method is shown to provide significant benefits as it explicitly incorporates state and control-dependent model uncertainty in safety analysis and, as a result, provides more robust, but not overly conservative controllers for the system. To summarize, the key contributions of this letter are two-fold: 1) We propose a framework for providing robust safety assurances to systems with unknown dynamics by incorporating state and control-dependent model uncertainty in HJ reachability analysis. We also provide a closed-form solution to the two-player zero-sum differential game between the control and disturbance player, in order to accommodate the state and control-dependent nature of the model uncertainty, and 2) We provide a concrete instantiation of our framework by modeling the system with an ensemble of neural networks and solving a robust optimal control problem to safeguard the learned model against bounded model uncertainty. 

\section{Problem Formulation}

In this work, we are interested in providing safety assurances for systems with unknown but deterministic \emph{control-affine} dynamics, as many practical systems are control-affine \cite{Lavalle06}, governed by ordinary differential equation 
\begin{equation}\label{eq:dyn}
    \frac{d\state}{d\tvar} = \dot{\state} = \dyn(\state,\ctrl) = \dyn_1(\state) + \dyn_2(\state)u
\end{equation}
where $\state \in \sset \subseteq \real^{\sdim}$ and $\ctrl \in \cset \subseteq \real^{\cdim}$ are the state and control of the system, and $\dyn_1(\state)$ and $\dyn_2(\state)$ are matrices of appropriate sizes. Given a set of states $\fset\subset\sset$, which we refer to as the \emph{failure set}, that the system must avoid (e.g., obstacles for a mobile robot), our goal is ensuring the system $\dyn$ does not enter the failure set $\fset$ for predefined time horizon $\thor$. More formally as stated in Prob. \ref{prob:safety_prob_general}.

\begin{problem}[Safety Problem for System $\dyn$]\label{prob:safety_prob_general}
    Given a failure set $\fset$ and time horizon $\thor$, obtain a safe set $\safeset\subseteq\sset$, along with a state-feedback controller $\pi$, such that the system $\dyn$ starting within $\safeset$ implies that $\dyn$ does not enter $\fset$ for $\thor$ using controls provided by $\pi$. A safe set is the maximal safe set, denoted by $\safeset^*$, if it contains all other safe sets as subsets. 
\end{problem}




Because we do not have access to the system dynamics $\dyn$, we further limit the scope of the problem to employing a model-based approach - first model the system of interest, then provide safety assurances on the system against any \emph{bounded} model uncertainty $\dstb\in\dset$, where $\dset$ is the set of admissible model uncertainty. We state the proposed safety assurance problem as follows: 


\begin{problem}[Safety Assurance under Model Uncertainty]\label{prob:safety}
    Given a failure set $\fset$ and time horizon $\thor$, obtain a set $\mathcal{R}\subseteq\sset$, along with a state-feedback controller $\pi_{\mathcal{R}}$, such that under the condition that all realized model uncertainties are contained within $\dset$, system $\dyn$ starts within $\mathcal{R}$ implies that $\dyn$ does not enter $\fset$ for $\thor$ using controls from $\pi_{\mathcal{R}}$, or equivalently, $\mathcal{R}\subseteq\safeset^*$. 
\end{problem}

\begin{definition}(Robust Safe Set)
    A solution to Problem 2, $\mathcal{R}\subseteq\sset$, is referred to as a robust safe set. A robust safe set is the maximal robust safe set, denoted as $\mathcal{R}^*$, is the robust safe set that contains any robust safe set as subsets.
\end{definition}

\section{Background}\label{sec:background}

In this section, we provide a brief overview of Hamilton-Jacobi (HJ) reachability analysis, an approach that can help solve Prob. \ref{prob:safety}. Let $g$ be a system described by dynamics $\dot{\state} = g(\state,\ctrl,\dstb)$, where $\state \in \sset \subseteq \real^{\sdim}$, $\ctrl \in \cset \subseteq \real^{\cdim}$, and $\dstb\in\dset\subseteq\real^{\ddim}$ are the state, control, and disturbance of the system. Disturbance $d$, in the interest of this letter, describes the model uncertainty. We use $\ctrlseq:\thor\rightarrow \cset$ and $\dstbseq:\thor\rightarrow \real^{\ddim}$ to denote the control and disturbance signals. Furthermore, we denote the state trajectory starting from state $\state$ at time $\tvar$ evolved with control and disturbance signals $\ctrlseq(\cdot)$ and $\dstbseq(\cdot)$ as $\traj_{\state,\tvar}^{\ctrlseq,\mathbf{\dstb}}$. With a slight abuse of the notation, we use $\traj_{\state,\tvar}^{\ctrlseq,\mathbf{\dstb}}(\tdummy)$ to denote the state at time $\tdummy\geq\tvar$ along the trajectory $\traj_{\state,\tvar}^{\ctrlseq,\mathbf{\dstb}}$.

Suppose $g(\state,\ctrl,\dstb)$ is uniformly continuous, bounded, and Lipschitz continuous in $\state$ for fixed $\ctrl$ and $\dstb$. We further assume that $\cset$ and $\dset$ are compact, and $\ctrlseq(\cdot)$ as well as $\dstbseq(\cdot)$ are measurable. Let $\tfunc(\state)$ be the signed distance function to $\fset$. We can obtain $\mathcal{R}^*$ by solving the following robust optimal control problem (Prob. \ref{prob:ind_safety}) with initial condition $\state = \state_0 \  \forall \state_0\in\sset$ and $t=0$. In robust control literature, Prob. \ref{prob:ind_safety} is posed as a two-player zero-sum differential game between the control $\ctrl$ and disturbance $\dstb$, who uses only nonanticipative strategies \cite{mitchell05}. Let $\Gamma(t)$ be the set of nonanticipative strategies, and $\mathbb{U}(t)$ be the set of admissible control signals. 

\begin{problem}[Robust Safety Optimal Control Problem] \label{prob:ind_safety}
\begin{equation*}
\begin{split}
    \inf_{\dstbseq(\cdot)\in\Gamma(t)} \sup_{\ctrlseq(\cdot)\in\mathbb{U}(t)}& \quad J(\state, \tvar, \ctrlseq, \dstbseq) = \min_{\tdummy \in [\tvar,\tend]} \tfunc(\traj_{\state,\tvar}^{\ctrlseq,\mathbf{\dstb}}(\tdummy)) \\
    s.t.     & \quad \dot{\state} = g(\state, \ctrl, \dstb)
    \end{split}
\end{equation*}
\end{problem}

\noindent Let us define the value function $\vfunc(\state,\tvar)$ to take on the optimal value of Prob. \ref{prob:ind_safety} at state $\state$ and time $\tvar$:
\begin{equation}\label{eq:vfunc}
    \begin{split}
    \vfunc(\state, \tvar) = & \inf_{\dstbseq(\cdot)\in\Gamma(t)} \sup_{\ctrlseq(\cdot)\in\mathbb{U}(t)} J(\state, \tvar, \ctrlseq, \dstbseq) \\
    &= \inf_{\dstbseq(\cdot)\in\Gamma(t)} \sup_{\ctrlseq(\cdot)\in\mathbb{U}(t)} \min_{\tdummy \in [\tvar,\tend]} \tfunc(\traj_{\state,\tvar}^{\ctrlseq,\mathbf{\dstb}}(\tdummy))
    \end{split}
\end{equation}

\noindent Then, $\mathcal{R}^*$ can be characterized using $\vfunc(\state,\tvar)$:
\begin{equation}
\mathcal{R}^* = \{\state \in \sset| \vfunc(\state,0) >0\}
\end{equation}

HJ reachability analysis provides a tractable means to compute the value function $V(\state,\tvar)$. It has been shown that $V(\state,\tvar)$ is the unique viscosity solution of the Hamilton-Jacobi-Isaacs Variational Inequality (HJI-VI) \cite{lygeros04,mitchell05}:
\begin{equation} \label{eq:HJIVI}
    \begin{aligned}
    \min \{&D_{\tvar} \vfunc + \ham(\state, \tvar, \nabla \vfunc), \tfunc(\state) - \vfunc(\state, \tvar) \} = 0 \\
    &\vfunc(\state, \tend) = \tfunc(\state), \quad \text{for} \ \tvar \in \left[0, \tend\right]
    \end{aligned}
\end{equation}
$\ham(\state, \tvar, \nabla \vfunc) = \max_{\ctrl}\min_{\dstb}\langle \nabla \vfunc, g(\state,\ctrl,\dstb)\rangle$ is the Hamiltonian. $D_{\tvar} \vfunc$ and $\nabla \vfunc$ denote the temporal derivative and the spatial gradients of $\vfunc(\state, \tvar)$. It is important to note that HJ reachability analysis also provides a state-feedback controller $\pi_{\mathcal{R}^*}$ that renders $\mathcal{R}^*$ control-invariant: 
\begin{equation}\label{eq:def_opt_ctrl}
    \pi_{\mathcal{R}^*}(\state,\tvar) = \arg\max_{\ctrl\in\cset} \min_{\dstb\in\dset} \langle \nabla \vfunc(\state,\tvar), g(\state,\ctrl,\dstb)\rangle
\end{equation}

\section{Safety Assurances for Learned Dynamics}\label{sec:method}

At the heart of our framework is solving a robust optimal control problem (Prob. \ref{prob:ind_safety}) to provide safety assurances for system $\dyn$, by incorporating the worst-case model uncertainty in the safety analysis. We obtain the maximal robust safe set $\mathcal{R}^*$ that safeguards the system against model uncertainty by solving the HJI-VI \eqref{eq:HJIVI} for $\vfunc(\state,\tvar)$. In this section, we first introduce the notion of \emph{uncertain model} and use it to set up the robust optimal control problem. Then, we show $\mathcal{R}^*$ does in fact confer safety assurance to $\dyn$. Finally, we present a concrete instantiation of our framework with $\dyn$ modeled by an ensemble of neural networks, and we discuss a method to quantify the model uncertainty from the ensemble.


\subsection{Model Representation and Hamiltonian Formulation}
Given a deterministic, continuous-time, control-affine system $\dyn(\state,\ctrl)$, we learn a nominal model $\bar{\dyn}(\state,\ctrl) = \bar{\dyn_1}(\state)+\bar{\dyn_2}(\state)\ctrl$ of $\dyn$. We assume there are \emph{bounded, state-dependent} model uncertainties $\dstb_1(\state)\in\dset_1(\state)\subseteq\real^{\sdim}$ and $\dstb_2(\state)\in\dset_2(\state)\subseteq\real^{\sdim\times\cdim}$, arising from errors of model approximation and unmodeled dynamics additive to $\bar{\dyn_1}(\state)$ and $\bar{\dyn_2}(\state)$. Hence, the \emph{uncertain} model $\hat{\dyn}(\state,\ctrl, \dstb_1, \dstb_2)$ can be written as 
\begin{equation}
    \hat{\dyn}(\state,\ctrl, \dstb_1, \dstb_2) = \bar{\dyn_1}(\state) +  \dstb_1(\state) +(\bar{\dyn_2}(\state)+\dstb_2(\state))\ctrl
\end{equation}
We refer to $\dset_1(\state)$ and $\dset_2(\state)$ as the \emph{model uncertainty bounds} on $\bar{\dyn}_1(\state)$ and $\bar{\dyn}_2(\state)$ at state $\state$, respectively. The Hamiltonian $\ham(\state, \tvar, \nabla \vfunc)$ formulated using $\hat{\dyn}(\state,\ctrl, \dstb_1, \dstb_2)$ is given in \eqref{eq:ham_uncertain_whole}, where $u^*, \dstb_1^*$, and $\dstb_2^*$ are the solutions to the maximin game in the RHS of \eqref{eq:ham_uncertain}.

\begin{subequations}\label{eq:ham_uncertain_whole}
    \begin{equation}\label{eq:ham_uncertain}
        \begin{split}
        \ham(\state, & \tvar, \nabla \vfunc(\state, \tvar)) = \max_{\ctrl\in\cset}\min_{\dstb_1\in\dset_1(\state)}\min_{\dstb_2\in\dset_2(\state)} \Bigl\langle \nabla \vfunc(\state, \tvar), \\
         & \bar{\dyn}_1(\state) + \dstb_1 + \left(\bar{\dyn}_2(\state) +\dstb_2 \right)\ctrl\Bigl\rangle \\
         \end{split}
    \end{equation}
    \begin{equation}\label{eq:ham_uncertain_sol}
        = \Bigl\langle \nabla \vfunc(\state, \tvar), \bar{\dyn}_1(\state) + \dstb_1^* + \left(\bar{\dyn}_2(\state) +\dstb_2^* \right)\ctrl^* \Bigl\rangle
    \end{equation}
\end{subequations}

The maximin game in \eqref{eq:ham_uncertain} does not generally have a closed-form solution since $\ctrl$ and $\dstb_2$ are multiplied together. We make the following assumptions to enable tractable computation of \eqref{eq:ham_uncertain}: 1) $\dset_1(\state)$ and $\dset_2(\state)$ are hypercubes containing the origin in their respective spaces $\real^{\sdim}$ and $\real^{\sdim\times\cdim}$, and 2) the set of admissible controls $\cset$ is a hypercube containing the origin in $\real^{\cdim}$. Under the assumptions, we can obtain closed-form solutions for $\ctrl$, $\dstb_1$ and $\dstb_2$.

\begin{note}
    The total model uncertainty is given by $\dstb(\state,\ctrl) = \dstb_1(\state) + \dstb_2(\state)\ctrl$, and hence the representation of $\dstb$ is state and control-dependent.
\end{note}

Let $\dstb_{1i}^*(\state)$ and $\nabla   \vfunc_i(\state,\tvar)$ denote the $i^{th}$ component of $\dstb_1^*(\state)\in\real^{\sdim}$ and $\nabla \vfunc(\state,\tvar)\in\real^{\sdim}$, respectively. Since $\dset_1(\state)$ is a hypercube containing the origin of $\real^{\sdim}$, we can write the uncertainty bound on the $i^{th}$ component of $\bar{\dyn}_1(\state)$ by an interval $\left[\underline{\dstb_{1i}(\state)}, \overline{\dstb_{1i}(\state)}\right]$. Then, $\dstb_1^*(\state)$ is given by 
\begin{equation}
    \dstb_{1i}(\state) = \begin{cases}
        \overline{\dstb_{1i}(\state)} \ \text{if} \ \nabla\vfunc_i(\state,\tvar) < 0 \\
        \underline{\dstb_{1i}(\state)} \ \text{if} \ \nabla\vfunc_i(\state,\tvar) \geq 0
    \end{cases}
\end{equation}

We now derive $\ctrl^*(\state)$, the optimal safety controller given in \eqref{eq:opt_ctrl}. Let us denote the $j^{th}$ component of $\ctrl^*\in\real^{\cdim}$ and the $j^{th}$ column of $\bar{\dyn_2}(\state)$ by $\ctrl_j^*$ and $\bar{\dyn}_{2j}(\state)$. Since $\cset$ is a hypercube in $\real^{\cdim}$, the bound on the $j^{th}$ component of $\ctrl$ is given by an interval $\left[\underline{\ctrl_j}, \overline{\ctrl_j}\right]$. Again, since $\dset_2(\state)$ is a hypercube in $\real^{\sdim\times\cdim}$, the model uncertainty bound on the $ij^{th}$ component of $\bar{\dyn}_2(\state)$ is an interval $\left[\underline{\dstb_{2ij}(\state)}, \overline{\dstb_{2ij}(\state)}\right]$. Let $\dstb_2^+(\state)\in\real^{\sdim\times\cdim}$ and $\dstb_2^-(\state)\in\real^{\sdim\times\cdim}$ be the ``best effort" $\dstb_2(\state)$ that intuitively try to decrease the Hamiltonian for positive or negative $\ctrl$, respectively. We denote the $j^{th}$ column of $\dstb_2^+(\state)$ and $\dstb_2^-(\state)$ by 
$\dstb_{2j}^+(\state)$ and $\dstb_{2j}^-(\state)$. More precisely, the $i^{th}$ component of $\dstb_{2j}^+(\state) \in \real^{\sdim}$ is given by $\overline{\dstb_{2ij}}$ when $\nabla \vfunc_i < 0$, $\underline{\dstb_{2ij}}$ when $\nabla \vfunc_i \geq 0$. The $i^{th}$ component of $\dstb_{2j}^-(\state) \in \real^{\sdim}$ is given by $\overline{\dstb_{2ij}}$ when $\nabla \vfunc_i \geq 0$, $\underline{\dstb_{2ij}}$ when $\nabla \vfunc_i < 0$. $\ctrl^*$ is then given in \eqref{eq:opt_ctrl}.

\begin{equation}\label{eq:opt_ctrl}
    \ctrl^*_j(\state) = \begin{cases}
        \vspace{0.1em}
        \overline{u_j},  \text{if} \  \nabla \vfunc(\state,\tvar)^\top \left(\bar{\dyn}_{2j}(\state) +  \dstb_{2j}^+(\state)\right) > 0\\ 
        \underline{u_j},   \text{if} \ \nabla \vfunc(\state,\tvar)^\top \left(\bar{\dyn}_{2j}(\state) + \dstb_{2j}^-(\state) \right) < 0  \\
        0, \text{otherwise}
    \end{cases}
\end{equation}

Lastly, we provide $\dstb_2^*(\state)$. Let us denote the $ij^{th}$ component of $\dstb_2^*(\state)\in\real^{\sdim\times\cdim}$ by $\dstb_{2ij}^*(\state)$. $\dstb_2^*(\state)$ is given below in \eqref{eq:opt_dstb_2}. 
\begin{equation}\label{eq:opt_dstb_2}
    \dstb_{2ij}^*(\state) = \begin{cases}
        \overline{\dstb_{2ij}(\state)} \ \text{if} \ \ctrl_j^*(\state)\nabla\vfunc_i(\state,\tvar) < 0\\
        \underline{\dstb_{2ij}(\state)} \ \text{if} \ \ctrl_j^*(\state)\nabla\vfunc_i(\state,\tvar) \geq 0
    \end{cases}
\end{equation}

We solve the HJI-VI \eqref{eq:HJIVI}, with the Hamiltonian \eqref{eq:ham_uncertain_sol} formulated for $\hat{\dyn}$, to obtain $\vfunc(\state,\tvar)$ and the maximal robust safe set $\mathcal{R}^*$ as well as its corresponding safety controller $\pi_{\mathcal{R}^*}$. Following directly from the definition of the value function \eqref{eq:vfunc}, $\mathcal{R}^*$ and $\pi_{\mathcal{R}^*}$ confer safety assurances to system $\dyn$, if for any state $\state\in\sset$, the realized model uncertainties at state $\state$ are contained within $\dset_1(\state)$ and $\dset_2(\state)$. This result is formalized in Lemma. \ref{thm:safety_assurance}.



\begin{lemma}\label{thm:safety_assurance}
    Given a control-affine system $\dyn(\state,\ctrl) = \dyn_1(\state) + \dyn_2(\state)\ctrl$ and its nominal model $\bar{\dyn}(\state,\ctrl) = \bar{\dyn}_1(\state) + \bar{\dyn}_2(\state)\ctrl$, let $\delta_1(\state) = \dyn_1(\state) - \bar{\dyn}_1(\state)$ and $\delta_2(\state) = \dyn_2(\state) - \bar{\dyn}_2(\state)$ be the realized model uncertainties. 
    If $\delta_1(\state)\in \dset_1(\state)$ and $\delta_2(\state)\in\dset_2(\state)$ $\forall \state\in\sset$, then $\mathcal{R}^*\subseteq\safeset^*$. 
\end{lemma}

\subsection{Ensemble Dynamics Representation}
Although our framework is agnostic to how the nominal model $\bar{\dyn}$ or the model uncertainties $\dstb_1$ and $\dstb_2$ are obtained, in this subsection, we introduce one method to jointly obtain $\bar{\dyn}$, $\dstb_1$, $\dstb_2$, and their corresponding bounds $\dset_1$ and $\dset_2$. 

In this work, we employ an ensemble of neural networks to obtain uncertainty in learned dynamics, a popular approach in literature to quantify uncertainty in deep networks \cite{deep_ensemble, chua18}. More specifically, given a deterministic, continuous-time, control-affine dynamical system $\dyn(\state,\ctrl)$, we learn an ensemble of $M$ fully connected feed-forward neural networks with control-affine architectures, and we use the ensemble as the nominal model $\bar{\dyn}$. More explicitly, the ensemble is given by
\vspace{-0.5em}
\begin{equation}
    E = \{NN^k(\state,\ctrl) = NN^k_1(\state)+NN^k_2(\state)u\}_{k=1}^{N_m}
\end{equation}
where $NN^k_1$ and $NN^k_2$ are neural networks.  With a slight abuse of notation, we denote the prediction of the ensemble $E$ by 
\vspace{-0.5em}
\begin{equation}\label{eq:nn_nom_model}
\begin{split}
    E(\state,\ctrl) &= \frac{1}{N_m}\sum_{k=1}^{N_m} NN^k(\state, \ctrl)\\
    &= \left(\frac{1}{N_m}\sum_{k=1}^{N_{\texttt{m}}} NN^k_1(\state)\right)+\left(\frac{1}{N_m}\sum_{k=1}^{N_m} NN^k_2(\state)\right)u\\
    &= \bar{\dyn}_1(\state) + \bar{\dyn}_2(\state) u = \bar{\dyn}(\state,\ctrl)
\end{split}
\end{equation}
\noindent We sometimes refer to $E(\state,\ctrl)$ as the \emph{mean dynamics}, as we are taking the mean prediction among the neural networks within the ensemble.


Given the setup of the nominal model $\bar{\dyn}(\state,\ctrl)$ in \eqref{eq:nn_nom_model}, we would like the model uncertainties $\dstb_1$ and $\dstb_2$ to intuitively quantify the \emph{variations of outputs} of the sub-nets $NN_1^k$ and $NN_2^k$ within the ensemble. We bound $\dstb_1(\state)$ and $\dstb_2(\state)$ using constant multiples of standard deviation of $\{NN^k_1(\state)\}_{k=1}^{N_m}$ and $\{NN^k_2(\state)\}_{k=1}^{N_m}$. Let us denote the extents of the $i^{th}$ and $ij^{th}$ dimensions of $\dset_1(\state)$ and $\dset_2(\state)$ by $\dset_{1i}(\state)$ and $\dset_{2ij}(\state)$, respectively. By assumption, $\dset_1(\state)\subseteq\real^{\sdim}$ and $\dset_2(\state)\subseteq\real^{\sdim\times\cdim}$ are hypercubes in their respective spaces, and therefore $\dset_{1i}(\state)$ and $\dset_{2ij}(\state)$ are real intervals. More precisely, the intervals are given by

\begin{subequations}
    \begin{equation}
        \dset_{1i}(\state) = [-\alpha\sigma_{1i}(\state), \alpha\sigma_{1i}(\state)]
    \end{equation}
    \begin{equation}
        \dset_{2ij}(\state) = [-\gamma\sigma_{2ij}(\state), \gamma\sigma_{2ij}(\state)] 
    \end{equation}
\end{subequations}

where $\sigma_{1i}(\state) = \texttt{StdDev}\left(\{NN^k_{1i}(\state)\}_{k=1}^{N_m}\right)$, $\sigma_{2ij}(\state) = \texttt{StdDev}\left(\{NN^k_{2ij}(\state)\}_{k=1}^{N_m}\right)$, $\texttt{StdDev}$ is a short hand for ``standard deviation", $NN^k_{1i}(\state)$ and $NN^k_{2ij}(\state)$ denote the $i^{th}$ and $ij^{th}$ outputs of the $k^{th}$ $NN_1$ and $NN_2$ sub-nets, respectively.

\begin{note}
    $\alpha$ and $\gamma$ are tunable parameters that determine conservativeness of the resulting safe set $\safeset$. As $\alpha$ and $\gamma$ increase, the model uncertainty bounds $\dset_1(\state)$ and $\dset_2(\state)$ increases $\forall \state\in\sset$. Accordingly, the safe set $\safeset$ shrinks. For all the experiments in this letter, we use $\alpha=\gamma=3$.
\end{note}



\section{Experiments}
\subsection{Inverted Pendulum}
In this example, we simulate an inverted pendulum with state $\state = [\theta, \dot{\theta}]^\top$ and control $\ctrl$, and its dynamics is given by $\dot{\state} = \left[\dot{\theta}, \ddot{\theta}\right]^\top = \left[\dot{\theta},  \frac{-b\dot{\theta} + \frac{1}{2}mgl\sin{\theta} - u}{\frac{ml^2}{3}}\right]^\top$
, where $l,m,g, $ and $b$ represents the length and mass of the pendulum, acceleration of gravity, and the friction coefficient, respectively. For the purposes of this case study, the analytical expression of the dynamics is assumed to be unknown.

We first train an ensemble dynamics model, consisting of 5 fully connected feed-forward neural networks with 3 hidden layers and 256 neurons per hidden layer, using dataset $\{(\state_i, \ctrl_i), \dot{\state_i}\}_{i=1}^M$ parsed from trajectory rollouts with random controls. In this experiment, we want the pendulum to avoid deviating from its unstable equilibrium for more than $0.6\pi$ radians. Equivalently, the failure set is given by $\fset = \{[\theta, \dot{\theta}]^\top|\theta>0.6\pi\} \cup \{[\theta, \dot{\theta}]^\top|\theta<-0.6\pi\}$.

Next, we compute the safe set $\safeset$, or equivalently complement of the backward reachable tube (BRT) of the failure set $\fset$, for a time horizon of 0.7 seconds, using the learned ensemble dynamics model along with model uncertainties. We also consider three baselines: 1) computing the safe set using only the mean dynamics $E(\state,\ctrl)$ (Baseline 1 Mean Dynamics), 2) computing the safe set with $\dset_1$ and $\dset_2$ calculated using split conformal prediction with 5000 calibration samples and marginal coverage of $95\%$ \cite{angelopoulos22_conformal_pred} (Baseline 2 Conformal Prediction), and 3) computing the safe set with $d_2(x)u$ approximated as $d_3(x)\in\real^{\sdim}$ (Baseline 3 Partial Game). The purpose of Baseline 3 is to remove the interaction of the control player $\ctrl$ and the disturbance player $\dstb_2$, and renders the model uncertainty \emph{control-independent}, as the action of the disturbance player $\dstb_3$ no longer depends on that of the control player $\ctrl$. Furthermore, the Hamiltonian computation \eqref{eq:ham_uncertain} decouples into three independent optimizations with respect to $\ctrl, \dstb_1$, and $\dstb_3$. The bound on the $i^{th}$ component of $\dstb_3$ is given by $\dset_3(x)_i = [-a(\state), a(\state)]$, where $a(\state) = \sum_{j=1}^{\cdim}\max\{|\underline{u_j}|, |\overline{u_j}|\} \times \max\{|\underline{d_{2ij}(\state)}|, |\overline{d_{2ij}(\state)}|\}$.

\begin{figure}
    \centering
    \includegraphics[width=0.8\linewidth]{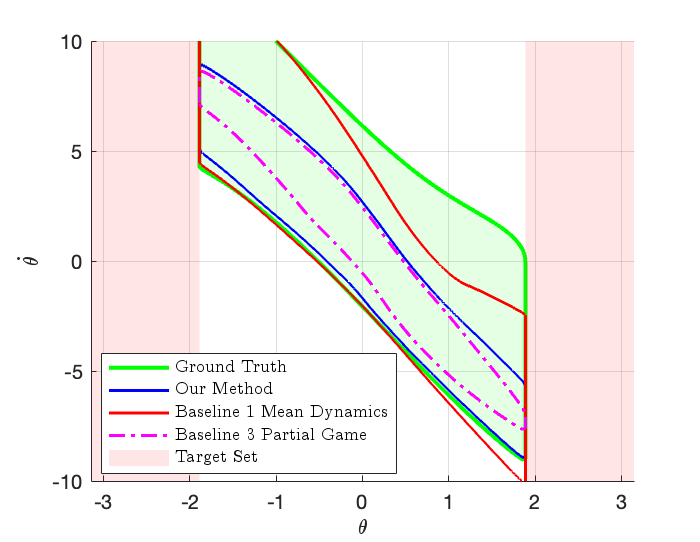}
    \vspace{-0.5em}
    \caption{Recovered safe sets (states starting from which the inverted pendulum stays within $0.6\pi$ of the upright for 0.7 seconds) for our method, the Mean Dynamics baseline (Baseline 1), the Partial Game baseline (Baseline 3) in the inverted pendulum experiment, with the ensemble trained with 300 training samples. Note that Baseline 2 (conformal prediction) is not visualized, because its safe set is empty.}
    \label{fig:inv_pend_brts}
\end{figure}

For the number of training samples $M=300$, we visualize the recovered safe sets for our method, all the baselines, and the ground truth in Fig. \ref{fig:inv_pend_brts}. The ground truth safe set, computed using the analytical expression of the dynamics, is shaded in green.
The safe set recovered using our method is entirely contained within the ground truth safe set, indicating satisfaction of the safety constraint. On the other hand, Baseline 1 fails to satisfy the safety constraint, since the recovered safe set is not contained within the ground truth safe set. The safe set from Baseline 2 is not visualized in Fig. \ref{fig:inv_pend_brts} because its recovered safe set is empty due to model uncertainty bounds $\dset_1$ and $\dset_2$ being too conservative. Specifically, conformal prediction provides \emph{state-independent} uncertainty bounds, which are dictated by the worst-case modeling errors across all states, leading to overly conservative behaviors. Baseline 3 is less conservative than Baseline 2, as its model uncertainty bounds are state-dependent, but it is more conservative than our methods since its model uncertainty $\dstb_3(\state)$ is control-independent and its bound $\dset_3(\state)$ is an overapproximation of that of our method.

\begin{figure}
    \centering
    \includegraphics[width=0.8\linewidth]{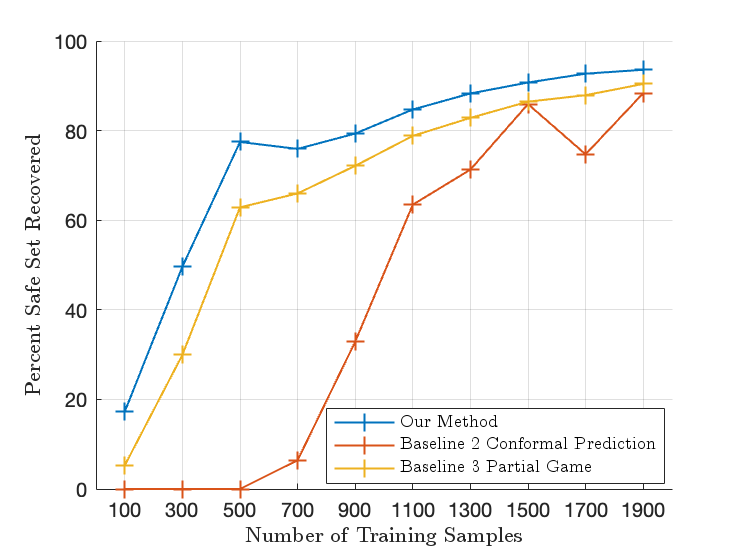}
    \caption{Changes in percent safe set recovered with our method, the Conformal Prediction baseline (Baseline 2), and the Partial Game baseline (Baseline 3), over the number of training samples.}
    \label{fig:inv_pend_ablation}
\end{figure}

We also perform an ablation study over the number of training samples $M$ to further highlight the benefit of using state and control-dependent model uncertainty. The percent safe set recovered, as a function of $M$, is charted in Fig. \ref{fig:inv_pend_ablation}. Across all experimented $M$, our method consistently outperforms the baselines, indicating that the state and control-dependent model uncertainty representation leads to less conservative behaviors. Our model uncertainty representation can reflect local variations of model uncertainties, allowing the safe set to expand or shrink according to local model uncertainty level. Furthermore, the control-dependent nature of our model uncertainty representation, which taken into consideration of the interaction between the control and the model uncertainty, also helps reduce the conservativeness.

\subsection{Turtlebot Hardware Experiment}

We apply the proposed approach on a real hardware testbed, TurtleBot 4, which we refer to as the \emph{vehicle}, in this experiment. We are interested in providing safety assurances for the vehicle \emph{carrying a payload}, and we seek to demonstrate the importance of incorporating model uncertainty in safety analysis and the benefit of model learning. 


Let $\mathcal{A}\subseteq\real^2$ be a rectangular experimental space, centered at $\left[0,0\right]^\top$ with side lengths $l_x$ and $l_y$. The failure set $\fset$ is hence given by $\fset = \{\state\in\sset||p_x|>\frac{l_x}{2}, |p_y|>\frac{l_y}{2}\}$, where $p_x$ and $p_y$ are the $x,y$ positions of the center of mass of the vehicle. We model the vehicle as a 4-dimensional system with state $\left[p_x, p_y, \theta,\omega\right]^\top$, where $\omega$ is the angular velocity, and control $\ctrl$, an angular velocity setpoint. We operate the vehicle with a constant forward velocity. 


We first collect 120 state and control trajectories using a random control at each time step. Each trajectory is roughly 5 seconds and yields about 100 training samples  $(\left[\state,\ctrl\right],\dot{\state})$. Then, we train an ensemble dynamics model containing 5 fully connected feed-forward neural networks, each with 3 hidden layers and 512 neurons per layer. 

\begin{figure}
    \centering
    \includegraphics[width=0.6\linewidth]{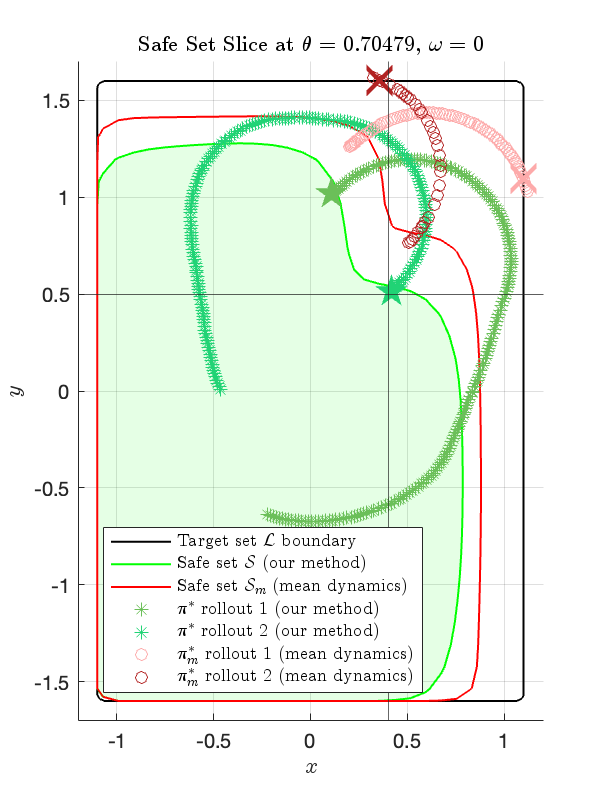}
    \vspace{-0.5em}
    \caption{Comparison between projected safe sets $\safeset$ and $\safeset_m$ along with TurtleBot rollout trajectories using safety controllers $\pi^*$ and $\pi^*_m$. The initial state of rollouts using $\pi^*$ are marked with pentagons. 
    All rollouts start with heading $\theta=0.7$ and angular velocity $\omega=0$. The states at which the TurtleBot enters failure set $\fset$ are marked with crosses.}
    \label{fig:tbot_vk_1}
\end{figure}

The safe set and the safety controller are computed for our method and the mean dynamics baseline until convergence (i.e., the time horizon is $\left[\tinit,\infty\right]$). For visualization purposes, we project the safe set to the $x,y$ plane at some fixed $\theta$ and $\omega$. In Fig. \ref{fig:tbot_vk_1}, the safe sets projected at $\theta=0.7$ and $\omega=0$ are shown. We apply mean dynamics baseline's safety controller $\pi^*_m$, from two states $\state_1 \ \text{and}\ \state_2$, within the mean dynamics baseline safe set $\safeset_m$. $\pi^*_m$ is unable to keep the vehicle inside $\mathcal{A}$, since $\safeset_m$ and $\pi^*_m$ do not take into consideration the model uncertainty and, as a result, are overly optimistic. On the other hand, we roll out the vehicle from 2 nearby states, $\state_3 \ \text{and}\ \state_4$, within our method's safe set $\safeset$ with our method's safety controller $\pi^*$, and the vehicle stays within $\mathcal{A}$, indicating that our method is able to obtain a better estimate of the actual safe set of the system.


We now highlight the benefit of model learning with a safety filtering experiment. When the vehicle is traveling at a constant forward velocity, it is common to model the vehicle with a three-dimensional Dubins Car (Dubins3D) with the dynamics $\left[\dot{p_x}, \dot{p_y}, \dot{\theta}\right]^\top = \left[V\cos(\theta), V\sin(\theta), \ctrl \right]^\top$. However, since the vehicle carries a payload, which introduces factors that could render the Dubins3D model inaccurate, a learned model can more accurately represent the system. We compute the safe set $\safeset_d$ and safety controller $\pi^*_d$ using the Dubins3D model to convergence. Then, we use $\pi^*_d$ to \emph{filter} a nominal controller $\pi(x) = 0$ in a least restrictive fashion \cite{borquez23}. The filtered controller $\hat{\pi}_d$ keeps the vehicle moving with the current heading unless it is at risk of exiting $\safeset_d$, in which case $\pi^*_d$ takes over (i.e. $\hat{\pi}_d(\state) = \pi^*_d(\state)$). We similarly filter $\pi$ with $\pi^*$, and filtered controller is denoted as $\hat{\pi}$. 

\begin{figure}
    \centering
    \includegraphics[width=0.6\linewidth]{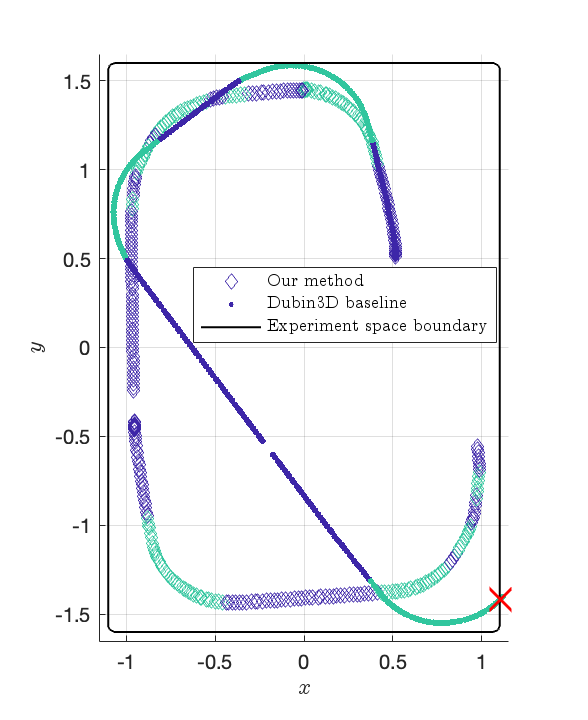}
    \vspace{-0.5em}
    \caption{Filtered controller $\hat{\pi}$ and $\hat{\pi}_d$ rollout trajectories. The blue markers indicate the states at which the nominal controller $\pi$ is on. Whereas the mint markers indicates the states where safety controllers $\pi^*$ and $\pi^*_{d}$ intervene. The state at which the vehicle exits the experiment space under $\hat{\pi}_d$ is marked with a red cross.}
    \label{fig:filtering_rollouts}
\end{figure}

Starting from a state $\state\in\safeset\cap\safeset_d$, we roll out the vehicle with $\hat{\pi}_d$ and $\hat{\pi}$. The state trajectories projected to the $x-y$ plane are shown in Fig. \ref{fig:filtering_rollouts}. The vehicle eventually exits $\mathcal{A}$ under $\hat{\pi}_d$, indicating that $\pi^*_d$ does not actually render $\safeset_d$ control invariant under the actual vehicle dynamics and is overly optimistic. However, $\hat{\pi}$ does keep the vehicle inside $\mathcal{A}$ for the entire experiment. For both trajectories, the mint-colored markers indicate the states where the safety controller $\pi^*$ or $\pi^*_d$ takes over and commands the vehicle to turn maximally to stay within $\mathcal{A}$. Since $\safeset_d$ is more optimistic than $\safeset$, $\pi^*_d$ intervenes closer to the boundary of $\mathcal{A}$ than $\pi^*$ would, and the resulting trajectory is uncomfortably close to exiting $\mathcal{A}$. In contrast, there is a healthy margin for the trajectory filtered by $\pi^*$.

\section{Conclusion}
In this letter, we proposed a framework for generating robust safety assurances for systems with unknown dynamics. Further, we provide a concrete instantiation of our framework with ensemble neural network models as the nominal model and safeguard the system against the worst-case model uncertainty. Though our method is shown to provide robust safety assurances in the experiments, it faces several challenges, which we look to address in future works. First of all, our method does not scale well to higher-dimensional systems. We will investigate the possibility of incorporating learning-based reachability computation tools \cite{bansal21dr,  fisac18_ham_rl} into our framework. Second, the proposed model uncertainty estimation approach might not provide model uncertainty bounds that accurately reflect the distribution of realized model uncertainties. We will address this challenge by examining other uncertainty estimation approaches, such as \cite{deep_ensemble} and \cite{gal15_dropout}.

\vspace{-0.5em}
\section*{ACKNOWLEDGMENT}
\vspace{-0.35em}
We would like to thank Albert Lin for insightful discussions during various phases of this project.

\bibliographystyle{plain}
\bibliography{citations} 

\end{document}

%% file: macros.tex
\newcommand{\state}{x}
\newcommand{\sset}{\mathcal{X}}
\newcommand{\sdim}{n_{\state}}
\newcommand{\ctrl}{u}
\newcommand{\ctrlseq}{\mathbf{\ctrl}}
\newcommand{\cset}{\mathcal{U}}
\newcommand{\cdim}{n_{\ctrl}}
\newcommand{\dstb}{d}
\newcommand{\dset}{\mathcal{D}}
\newcommand{\dstbseq}{\mathbf{\dstb}}
\newcommand{\ddim}{n_{\dstb}}
\newcommand{\fset}{\mathcal{F}}
\newcommand{\tfunc}{l}

\newcommand{\safeset}{\mathcal{S}}
\newcommand{\ham}{H}
\newcommand{\vfunc}{V}

\newcommand{\traj}{\xi}

\newcommand{\tvar}{t}
\newcommand{\tend}{T}
\newcommand{\tinit}{0}
\newcommand{\tdummy}{\tau}
\newcommand{\thor}{[\tinit,\tend]}
\newcommand{\dyn}{f}

\newcommand{\real}{\mathbb{R}}
\newtheorem{problem}{Problem}

\newtheorem{lemma}{Lemma}
\newtheorem{definition}{Definition}
\newtheorem{note}{Note}